\documentclass{article}

\usepackage{microtype}
\usepackage{graphicx}
\usepackage{booktabs} 
\usepackage{makecell}
\usepackage[T1]{fontenc}
\usepackage{listings}
\usepackage{color}
\usepackage{filecontents}       
\usepackage{caption}
\usepackage{subcaption}
\usepackage[utf8]{inputenc}
\usepackage{amsmath}
\usepackage[nopar]{lipsum}

\definecolor{dkgreen}{rgb}{0,0.6,0}
\definecolor{gray}{rgb}{0.5,0.5,0.5}
\definecolor{mauve}{rgb}{0.58,0,0.82}

\usepackage{hyperref}

\usepackage[accepted]{icml2020}

\begin{document}

\twocolumn[
\icmltitle{Multi-scale Transformer Language Models}




\begin{icmlauthorlist}
\icmlauthor{Sandeep Subramanian}{1,2}
\icmlauthor{Ronan Collobert}{1}
\icmlauthor{Marc'Aurelio Ranzato}{1}
\icmlauthor{Y-Lan Boureau}{1}
\end{icmlauthorlist}

\icmlaffiliation{1}{Facebook AI Research}
\icmlaffiliation{2}{Mila, Université de Montréal}

\icmlcorrespondingauthor{Sandeep Subramanian}{sandeep.subramanian.1@umontreal.ca}

\icmlkeywords{Language Modeling, Multi-scale, Transformer}

\vskip 0.3in
]



\printAffiliationsAndNotice{\icmlEqualContribution} 

\begin{abstract}
We investigate multi-scale transformer language models that learn representations of text at multiple scales,
and present three different architectures that have an inductive bias to handle the hierarchical nature of language. Experiments on large-scale language modeling benchmarks  empirically demonstrate favorable likelihood vs memory footprint trade-offs, e.g. we show that it is possible to train a hierarchical variant with 30 layers that has 23\% smaller memory footprint and better perplexity, compared to a vanilla transformer with less than half the number of layers, on the Toronto BookCorpus. We  analyze the advantages of learned representations at multiple scales in terms of memory footprint, compute time, and perplexity, which are particularly appealing given the quadratic scaling of transformers' run time and memory usage with respect to sequence length.
\end{abstract}

\section{Introduction}

Human language displays simultaneous organization at multiple scales and granularities: topics are maintained over long spans of text without controlling every single word, while grammatical correctness imposes strong local constraints without influencing word choice at a very long range. The choice of a word is thus constrained by both local and longer-range information.
This multi-scale structure is reminiscent of well-studied hierarchies in many domains of natural perception. In vision, the resolution of receptive fields decreases as one moves either away from the center of the retina, or higher in the hierarchy of visual cortex areas (from V1 to V2 to V4).
This gives rise to phenomena such as crowding \citep{pelli2008uncrowded}, where only some ensemble statistic of a set of elements can be perceived, and the existence of metamers, distinct visual stimuli that look the same when seen at the periphery \citep{freeman2011metamers,wallis2019image}. Incorporating both local and global organizational constraints and compressing representations with statistical pooling have been suggested to better explain empirical perceptual vision phenomena \citep{wallis2019image}. Multi-resolution pyramids such as Laplacian pyramids have been shown to produce more convincing image generation \citep{denton2015deep} and efficient compression \citep{burt1983laplacian}. Combining representations at multiple scales has also been successful for speech and audio generation \citep{mehri2016samplernn}.

Language modeling efforts however, typically rely on modeling the multi-scale nature of language without strong architectural priors. Representations are learned only at the finest scale, usually that of words or subwords, and rely on the training objective of predicting the next word in the sequence to implicitly capture the need to maintain consistency and coherence across multiple scales. Some previous efforts have advocated for multi-scale sequence models \cite{koutnik2014clockwork,chung2016hierarchical,mehri2016samplernn}, but these haven't seen widespread adoption in language modeling on current large-scale benchmarks.
The potential of leveraging multiple scales of representation to reduce 
memory footprint (see Section \ref{sec:space_time_complexity}) is especially appealing for language modeling because  transformers~\citep{DBLP:journals/corr/VaswaniSPUJGKP17}, which are 
currently the most popular architectures, 
suffer from quadratic memory usage scaling as context length increases.

We argue here that multi-scale transformer architectures can lead to better and more efficient generative models of text. We design three such architectures and present empirical evidence of their advantages, as well as analyses to better understand how transformer language models (LMs) use available context, following work by \citet{khandelwal2018sharp,sankar2019neural}. 
Our work thus makes the following contributions:
    (1) We present three different multi-scale transformer architectures for language modeling.
    (2) We show that on some benchmarks, these models typically have smaller memory footprint for the same performance. For example, we show that it is possible to train a multi-scale variant with 30 layers that has 23\% smaller memory footprint and better perplexity, compared to a vanilla transformer with less than half the number of layers.
    (3) We show that transformer LMs suffer only minor perplexity increases when only looking at the last 8 or 16 tokens instead of 512, and show that combining a short context of 8 tokens at the finest granularity with longer contexts at a coarser scale has perplexity similar to that of a model that looks at the entire longer context at the finest granularity.

\section{Related Work}
Transformer-based language models \cite{DBLP:journals/corr/VaswaniSPUJGKP17,radford2018improving,al2019character} have become the model of choice for most large-scale language modeling benchmarks. \citet{kaplan2020scaling} report power-law scaling of transformer language models with model capacity and data size. As models get bigger, the amount of computational resources, especially memory, grows quickly. In the next paragraph we review some recent efforts attempting to address this issue.

\paragraph{Memory-Efficient Transformers.}
\citet{sukhbaatar2019adaptive} present an adaptive attention mechanism that
learns how far back into the past each head in a transformer should look, and if implemented efficiently with sparse matrix operations, can help save memory. \citet{liu2018generating,rae2019compressive} present approaches that compress the transformer's memory with strided convolutions. Specifically, \citet{liu2018generating} compress the keys and values in the multi-headed attention by a factor of 3 for long-text abstractive summarization. \citet{child2019generating} present sparse transformers along with efficient CUDA kernels for sparse attention demonstrating the ability to generate very long sequences. \citet{rae2019compressive} compress the recurrent memory for a transformer-XL \cite{dai2019transformer}, but find that the best performing variant is one that does not learn the compression function end-to-end. \citet{liu2019hierarchical} proposes a hierarchical extension of the architecture proposed in \citet{liu2018generating} that attends over very long sequences, with the aim to better model paragraph- and document-level contexts. \citet{kitaev2019reformer} describe the different factors that contribute to large memory footprints in vanilla transformers. They use reversible layers to remove the need to store activations at every layer in the forward pass, LSH attention to decrease memory requirements from $\mathcal{O}{(N^2)}$ to $\mathcal{O}{(N\log{}N)}$ where $N$ is the sequence length, and they split activations in the feedforward layers. \citet{bai2019deep} present a way to train infinite-depth weight-tied feedforward nets via root-finding, yielding  a memory footprint that is constant with the depth of the network, but incurs computational overhead, similar to reversible layers. Unfortunately, this computational overhead is typically at least 50\% \cite{gomez2017reversible}.

\paragraph{Multi-scale architectures} for sequential data \cite{schmidhuber1992learning, el1996hierarchical, koutnik2014clockwork} have attempted to exploit the multi-scale nature of data like language, speech, music and audio. Such models seek to build representations of the input signal at multiple resolutions
to better control the generative process by keeping representations of the high-level structure in the signal (e.g., topic) invariant to fine-resolution local changes (e.g., precise word choice or grammar), while simultaneously allowing flexibility at finer resolutions to model these complex phenomena. They have seen success in modeling conversations \cite{sordoni2015hierarchical}, raw audio, speech and music \cite{mehri2016samplernn}, and language \cite{chung2016hierarchical}. \citet{sordoni2015hierarchical} use the available hierarchical structure in conversations such as utterance level boundaries to define a shallow hierarchy. \citet{koutnik2014clockwork} and \citet{mehri2016samplernn} use fixed time scales, while \citet{chung2016hierarchical} try to learn the clocking function using the task-specific training objective in an end-to-end manner, with an RNN-based architecture. \citet{garg2019multiresolution} present multi-resolution transformers where resolution hierarchies are obtained from explicit boundaries in the data such as sentences or paragraphs. This however limits the model's ability to capture hierarchies that aren't explicit in the data, e.g. within a single sentence or across sentences within the same paragraph. To the best of our knowledge, more generic multi-scale architectures not relying on limited data-dependent boundaries haven't been explored in conjunction with transformer language models.

\begin{figure*}[ht!]
\centering
\begin{subfigure}{.5\textwidth}
  \centering
  \includegraphics[width=0.8\textwidth,scale=0.7]{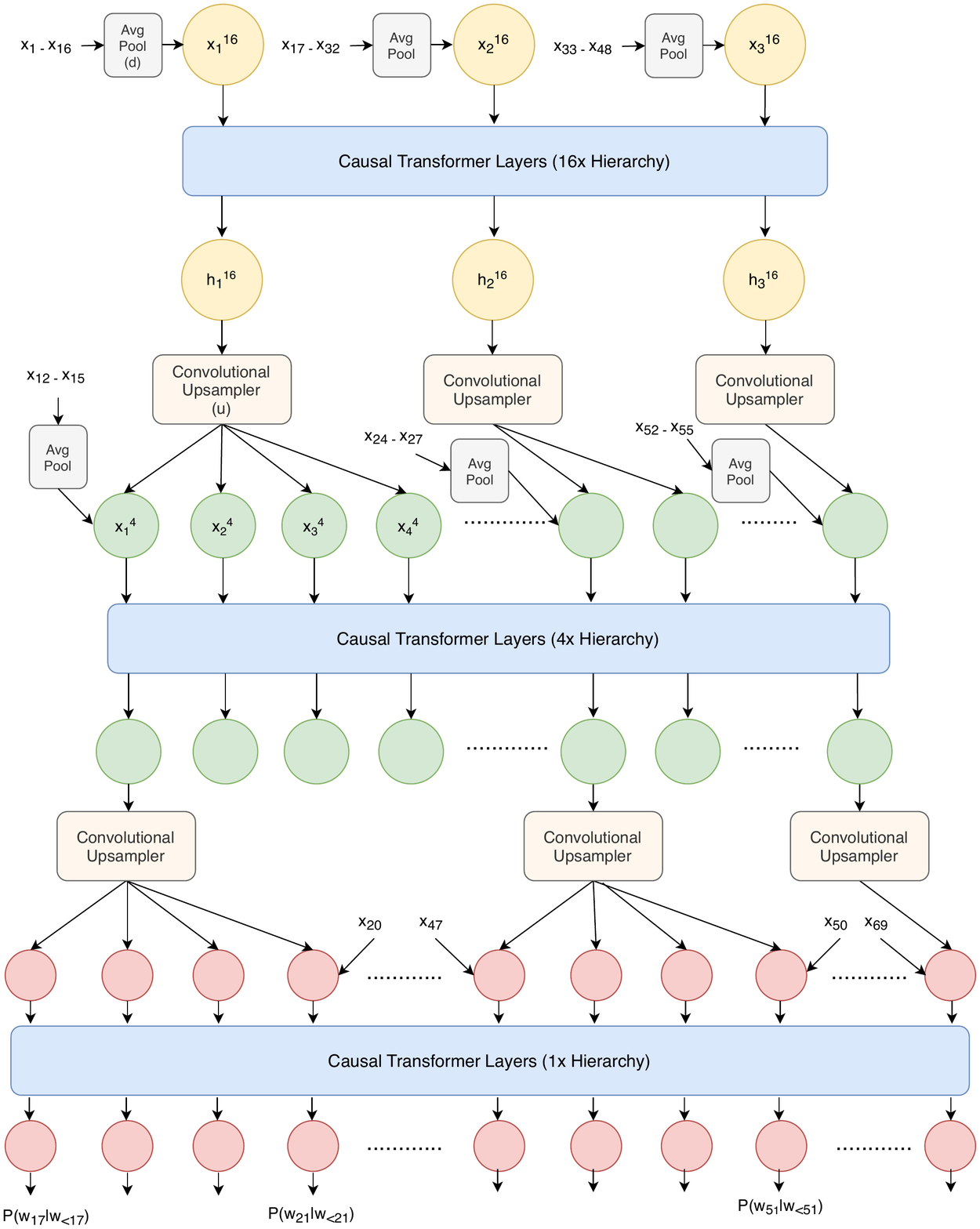}
  \caption{Top-down Model}
  \label{fig:topdown}
\end{subfigure}%
\hspace{-4em}
\begin{subfigure}{.5\textwidth}
  \centering
  \includegraphics[width=0.75\textwidth,scale=0.6]{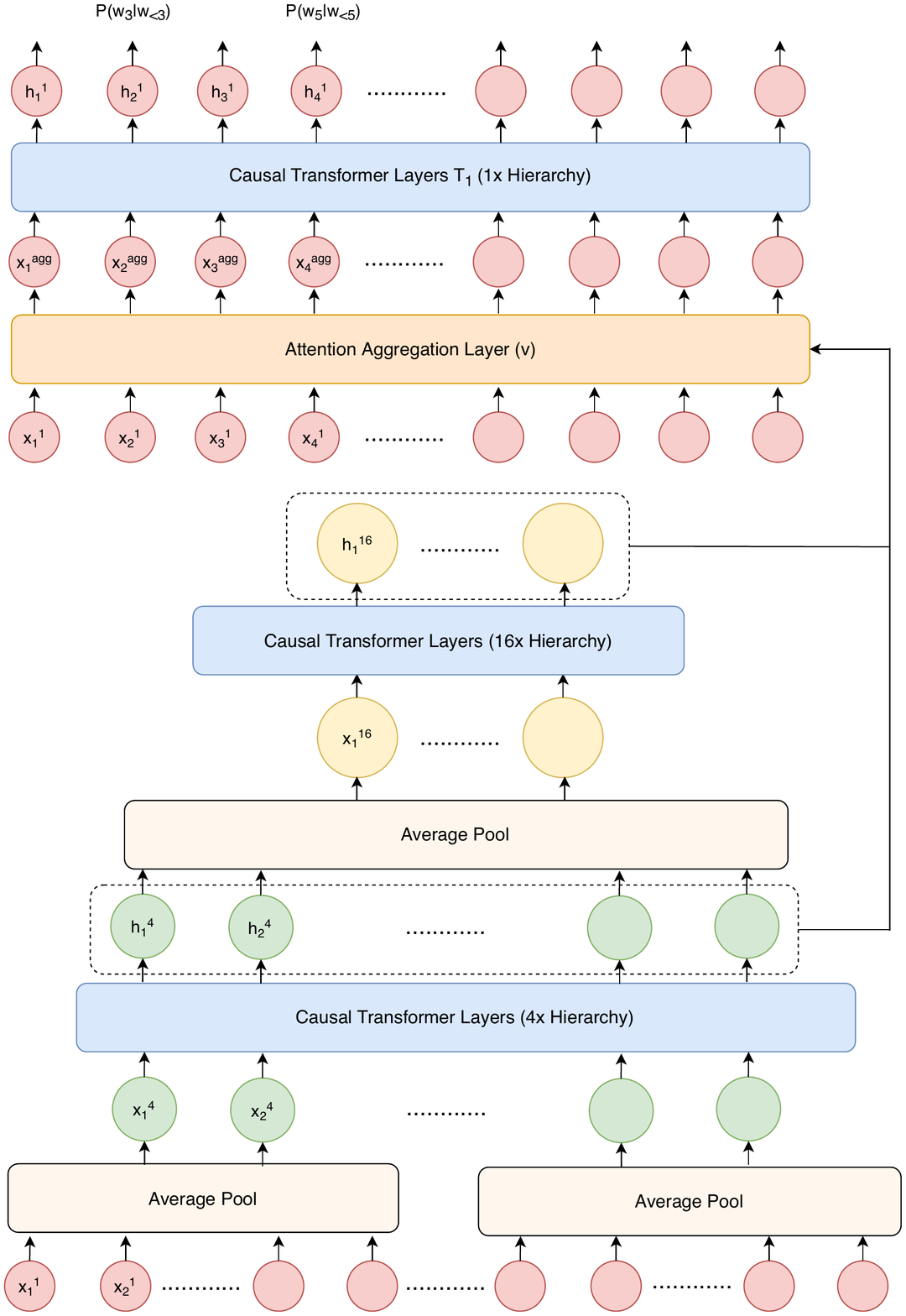}
  \caption{Bottom-up Model}
  \label{fig:bottomup}
\end{subfigure}%
\hfill
\begin{subfigure}{.3\textwidth}
  \centering
  \includegraphics[width=0.69\textwidth,scale=0.2]{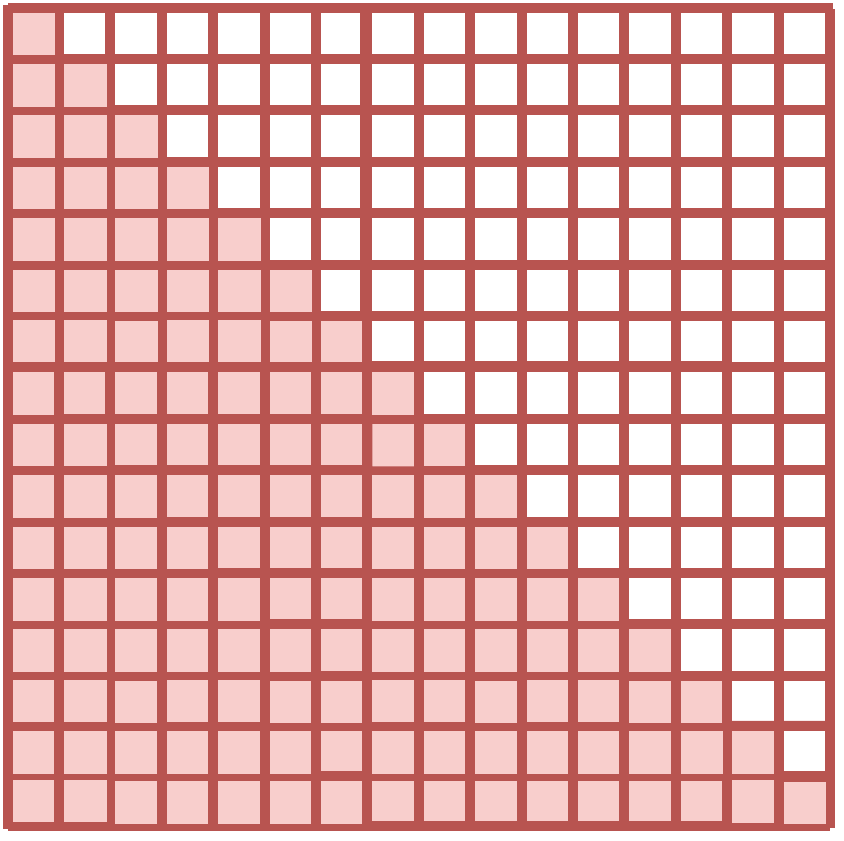}
  \caption{Vanilla attention mask}
  \label{fig:vanilla_attn_mask}
\end{subfigure}%
\hspace{-4em}
\begin{subfigure}{.3\textwidth}
  \centering
  \includegraphics[width=0.69\textwidth,scale=0.2]{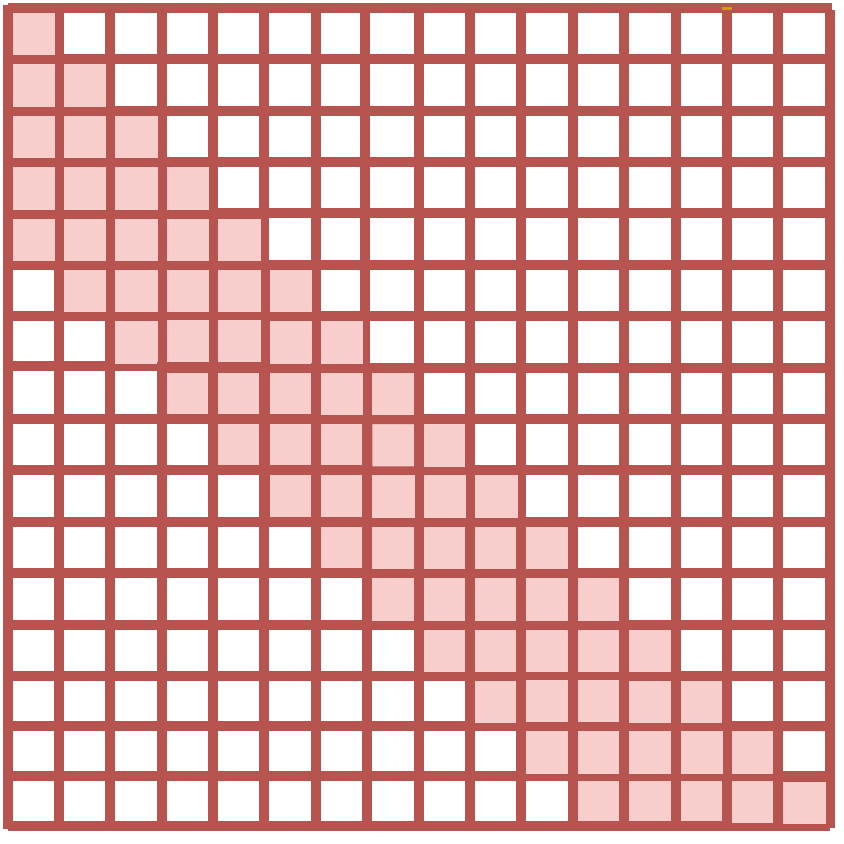}
  \caption{Local attention mask}
  \label{fig:local_attn_mask}
\end{subfigure}%
\begin{subfigure}{.37\textwidth}
  \centering
  \hspace{-3em}
  \vspace{-0.5em}
  \includegraphics[width=\textwidth,scale=0.4]{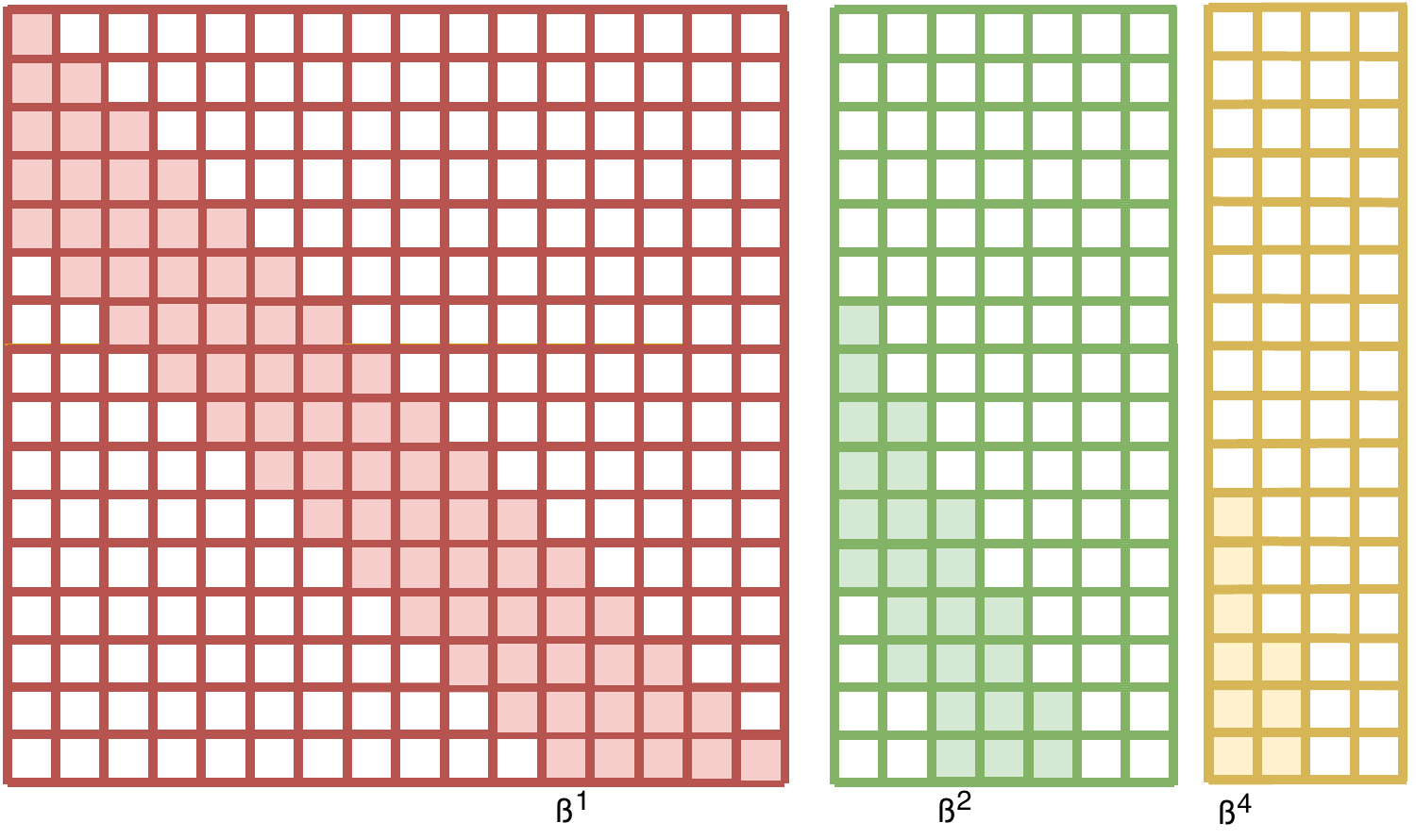}
  \caption{Retina attention mask}
  \label{fig:retina_attn_mask}
\end{subfigure}%
\caption{Our proposed model architectures: (a) \textbf{Top-down model} that builds representations from coarse (yellow) to fine (red) scales. (b) \textbf{Bottom-up Model} that does the opposite while aggregating representations from different scales before operating at the finest scale. (c,d,e) \textbf{Retina Model} that treats different parts of of the model's context with different granularities - nearby information is fine-grained, but coarse in the distant past. We visualize this using the model's attention mask. (c) is a standard autoregressive mask in vanilla transformers, (d) a local attention mask \cite{parmar2018image} that only looks at local information (e) our proposed retina variant with green and yellow indicating progressively coarser scales}
\label{fig:model_archs}
\end{figure*}

\section{Models}
We first briefly review language modeling and transformers \cite{DBLP:journals/corr/VaswaniSPUJGKP17} which are the base for our architectures.
We then describe the three different multi-scale architectures we explore for language modeling. See Figure \ref{fig:model_archs} for a visual depiction of our proposed architectures and Listings 1, 2 and 3 in Appendix for a PyTorch-like implementation of the forward pass through the models.

\subsection{Language Modeling \& Transformer Preliminaries}
The goal of language modeling is to estimate the joint probability of a sequence of words (or subwords, but we will refer to ``words'' for simplicity). The models we consider in this work factorize the joint probability over words into a product of conditional probabilities of each word given everything that precedes it. Conditional probabilities are typically parameterized as recurrent neural networks like LSTMs \cite{graves2013generating, mikolov2010recurrent} or feedforward models like gated convnets \cite{dauphin2017language}, transformers \cite{DBLP:journals/corr/VaswaniSPUJGKP17, radford2018improving}, or estimated with non-parametric count-based statistics like in n-gram language models.

Transformers have
become ubiquitous for large-scale language modeling \cite{radford2018improving,baevski2018adaptive,radford2019language,dai2019transformer,kaplan2020scaling}. They are self-attentive models that have stacks of residual blocks, each of which contains layers of multi-headed self-attention and feedforward modules. Multi-headed attention is a generalization of dot-product attention \cite{bahdanau2014neural, luong2015effective} where a score is computed between a query $Q$ and key $K$ for different learned projections of both. The scores are then normalized with a softmax and used as weights to compute a weighted average over values $V$ at each position in the sequence. In language models, we use self-attention where $Q=K=V$. The resulting representations are then summed with the input to the residual block and then normalized using LayerNorm \cite{ba2016layer} followed by a feedforward layer, normalization, and residual summing again.
All operations in the model have the advantage of being easily parallelizable on current hardware, and the attention mechanism
allows the model to learn long-range dependencies.

In the rest of this section, we  consider estimating the joint probability of $n$ words, factorized autoregressively.
We 
use $d_{model}$ and $d_{ff}$ to refer to the dimensionality of the intermediate hidden states 
and the convolutional feedforward layer in the model, respectively.
Words are first represented as embeddings $x_1 \ldots x_n$ of dimension $d_{model}.$
Let $k_1 < \ldots < k_m,$  denote the $m$ scales of a multi-scale model, where $k_1 = 1$ is
the finest scale that looks at every token, and scales $k_i$ incorporates information that has been down-sampled by a factor $k_i$ through a down-sampling operator $d$ (e.g., an average over windows of size $k_i$ or a strided convolution). For example,
a model with scales $(k_1, k_2, k_3) = (1, 4, 16)$ would use three scales
with each scale being 4 times coarser than the preceding scale.
Let $x^{k_i}$ denote the representation that is fed as input to vanilla transformer blocks $t_{k_i}$ at scale $k_i$ (with possibly different numbers of layers at each scale), and $h^{k_i}$ the representation at scale $k_i$. We now detail how $x^{k_i},h^{k_i}$ are computed for each
of the variants we propose.

\subsection{Top-down Model}
\label{sec:topdown}
The Top-down model  we propose (Fig.~\ref{fig:topdown}) is largely inspired by SampleRNN \cite{mehri2016samplernn}, a multi-scale recurrent architecture for generating audio. We call this a ``Top-down'' model because it builds representations of the sequence progressively from coarser to finer scales, by running multiple transformer layers at a particular coarse scale followed by convolutional upsampling to the subsequent (finer) scale. Predictions are made at the finest scale. 

The Top-down transformer uses downsampling operators $d$, that take as input a sequence of vectors and a factor $k_i$ by which to downsample them, to compute a representation of the input at scale $k_i$ of length of $n/k_i$; in our experiments, $d$ is average pooling with a kernel of size $k_i$, or causal strided convolutions.

The input to the coarsest scale in the model $x^{k_m}$, is simply the pooled token embeddings over windows of size $k_m$. Inputs to the finer-scale transformers thereafter are obtained by combining the pooled token embeddings at the corresponding scale with the upsampled representation from the immediately coarser scale:

\noindent
\begin{eqnarray*}
\bar{x}^{k_i} &=& d(x_1 \ldots x_n, k_i), m \geq i \geq 1, \\
x^{k_m}  &=& \bar{x}^{k_m},\\
h^{k_i} &=& t_{k_i}(x^{k_i}), m \geq i \geq 1, \\
x^{k_{i-1}} &=& f(\bar{x}^{k_{i-1}}, u(h^{k_i}, k_i/k_{i-1})),  m \geq i \geq 1,
\end{eqnarray*}
\noindent

where $f$ denotes a function that concatenates its inputs (of equal dimension), followed by a learned linear projection to half the dimension of the concatenated vector; $u$ is an upsampling function such that 
$u(h^{k_i}, k_i/k_{i-1})$ 
indicates that representations 
$h^{k_i}$ 
are being upsampled by a factor $k_i/k_{i-1}$.
The model is trained to predict the next word in the sequence using representations $h^{1}$. To make sure representations aren't informed by the future at any position, we slice and shift the inputs to $d$ appropriately (see Listing 1 and Figure \ref{fig:topdown}).

The motivation behind such a model is to have early layers learn high-level or coarse outlines of what the model should be generating and have those representations be progressively upsampled to include finer and finer details. Vanilla transformer language models by contrast, entangle the learning and representation of coarse and fine details.

A benefit of this model over vanilla transformer LMs is that they are typically much faster at inference. This is because transformer layers at the coarser scales do not need to run at every time step, but only once every $k_i$ steps for a particular scale. A Top-down model with 26 layers is about 30\% faster than a vanilla transformer with 14 layers (see Appendix Table \ref{tab:sample_level_eval}).

\subsection{Bottom-up Model}
In contrast to the Top-down model, the Bottom-up model (Fig.~\ref{fig:bottomup}) builds representations progressively from fine to coarse scales. 
Instead of upsampling operators, the architecture uses an aggregation layer, denoted by $v$, 
to incorporate information from coarser scales at the word level.  
This is a transformer layer where certain subsets of heads attend to representations from different scales. Specifically, inputs to the multi-headed attention module within the layer are the word embeddings themselves denoted by $x_1 \ldots x_n$, and the keys and values are representations at different coarser scales denoted by $h^{k_2} \ldots h^{k_m}$; we denote this operation via $v(h^{k_2} \ldots h^{k_m}, x_1 \ldots x_n)$.
This aggregation layer is given as input to the finest transformer blocks that make word-level predictions. The attention mask is constructed appropriately for each subset of heads to prevent looking at the future. To summarize:
\noindent
\begin{eqnarray*}
h^{k_1} & = & h^1 = x_1 \ldots x_n, \\
x^{k_i} & = & d(h^{k_{i-1}},k_i/k_{i-1}), 2 \leq i \leq m,\\
h^{k_i} & = & t_{k_i}(x^{k_i}), 2 \leq i \leq m, \\
x^{agg} & = & v(h^{k_2} \ldots h^{k_m}, x_1 \ldots x_n),\\
h^{out} & = & t_1(x^{agg}),
\end{eqnarray*}
\noindent
where $h^{out}$ is the output from which word probabilities are predicted.

\subsection{Retina Model}
We also experiment with an architecture that combines progressively coarser representations for tokens further away in the past, in a way that is reminiscent of the progressively bigger receptive fields in the retina as one moves away from the center. The underlying intuition is that a lot of the fine-grained information at the word level might be necessary in a close range (e.g., for grammaticality), but not so much at larger distances (see also the shuffling results in Section~\ref{sec:ablation}).
Unlike the Top-down and Bottom-up models, this \textit{Retina} model does not have separate transformer layers at every scale. The Retina model looks at each token with a different granularity: tokens in the recent past are looked at with finer granularity and distant tokens at a coarser granularity.
This is a simple modification to the multi-headed attention module of a transformer: namely, a different subset of heads look at representations from different scales. The queries to the multi-headed attention always come from the representations at the finest 1x ($k_1$) scale, while the keys and values are selected appropriately based on each input query position and the scale 
assigned to each head. 
This amounts to having local attention \cite{parmar2018image} at the finest 1x ($k_1$) scale (for the first subset of heads), and \textit{memory compressed attention} \cite{schmidhuber1992learning,liu2018generating, rae2019compressive} with added sparsity for the remaining coarser scales.
Note that context windows across the attention heads at different scales can have arbitrary boundaries for each scale (see Fig.~\ref{fig:retina_attn_mask}).

Formally, consider the same setting as in the Top-down model (see Sec.~\ref{sec:topdown}), with scales $k_1, \ldots, k_m$ and token embeddings $x_1, \ldots, x_n$, predicting token at position $t.$
Downsampled representations $\bar{x}^{k_i}$ are obtained at various scales through average pooling or strided convolutions, and then used as keys and values in a transformer. 
Each scale uses its own context window. These context windows are obtained by slicing the full-length context into non-overlapping windows of sizes roughly proportional to the $k_i$ scales, with the coarser scales using larger, further away context windows. Denoting context window boundaries by $\beta^{k_i}$ and assuming a fixed scale ratio $r=k_i/k_{i-1},$ this means that
$\beta^{k_i}$ roughly grows geometrically as $r^i,$ and each scale takes up about the same space for representation,
since the context size increase exactly compensates the higher downsampling ratio.
Overall, the output representation $h^{out}$ is given by:

\noindent
\begin{eqnarray*}
    \bar{x}^{k_i} &=& d(x_{t-\beta^{k_i}} \ldots x_{t-\beta^{k_{i-1}}-1}, k_i), m \geq i \geq 1, \\
    h^{out} &=& t(\bar{x}^{k_1} \ldots \bar{x}^{k_m}). 
\end{eqnarray*}
\noindent

This basic description can easily be extended to arbitrary non-overlapping context window boundaries (see ablation in Section~\ref{sec:ablation}), while still maintaining
the general design that the context windows of coarser scales are bigger and further away from $t.$

\subsection{Space \& Time Complexity for Transformer Layers}
\label{sec:space_time_complexity}
Transformer language models typically require large amounts of GPU memory to train. There are a few factors that contribute to this. 

\begin{enumerate}
    \item They require intermediate activations at every time step and layer to be stored in memory for fast backpropagation. Memory footprint therefore scales linearly with the number of layers and we'd often like to fit as big a model as we can into memory \cite{kaplan2020scaling}.
    \item Inside a transformer layer, multi-headed attention with queries, keys and values $Q,K,V$ requires computing a score for every pair of elements in the query and key (i.e.) the $QK^T$ matrix is of size $N\times N$ (assuming a batch size of 1) where $N$ is the number of elements in the sequence. Both memory and time complexities are therefore quadratic in the length of the sequence $\mathcal{O}(N^2)$, computation however is typically less of a problem since it can be parallelized easily across $N$. With short sequences, where the embedding dimension of the queries, keys, and values are greater than the sequence length, the $(QK^T)V$ matrix becomes expensive to store and compute. The overall complexity is $\mathcal{O}(N^2 + NH)$, where $H$ is the embedding dimension.
    \item Transformer layers also contain position-wise linear layers that are about 4 times as wide as the model embedding dimension and therefore require 4 times as much memory to store activations. In typical language modeling setups, where $N \approx H$, these activations dominate memory footprint (see Appendix Table \ref{tab:memory_footprint}).
\end{enumerate}

\begin{figure}[t]
    \center{\includegraphics[scale=0.35]
    {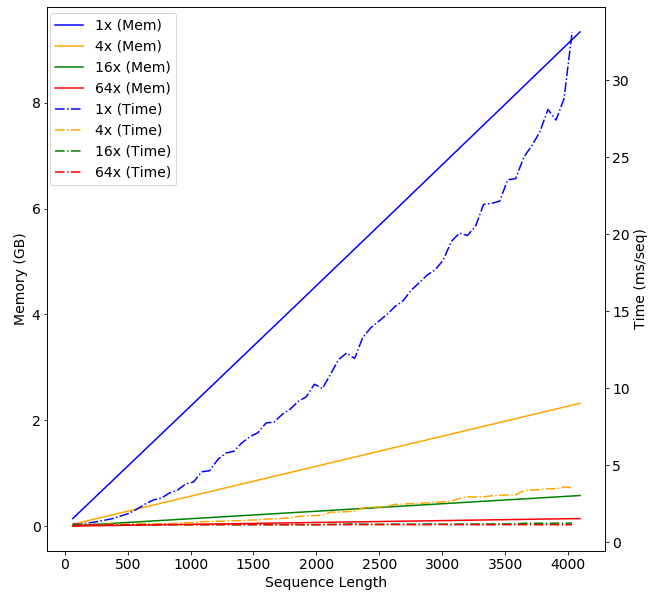}}
    \caption{\label{fig:seq_len_vs_mem_scales} Memory footprint and run time vs sequence length and run time for a single transformer LM layer at different scales}
\end{figure}

Our multi-scale architecture addresses 1 and 2 by reducing the number of positions over which transformer layers at coarser scales need to operate on. Figure \ref{fig:seq_len_vs_mem_scales} shows memory footprint in GB and time taken for a forward pass in milliseconds for a single transformer layer at different scales as a function of the input sequence length. Each transformer layer in this setting is identical to the ones used in our BookCorpus and Wikitext-103 experiments. Layers at the 4x, 16x and 64x scales are more efficient in terms of time and memory. Memory footprint was calculated analytically by accounting for the number of stored activations in the layer and time by implementing the layer in PyTorch.

By contrast, recurrent models have far fewer layers (typically an order of magnitude smaller) and are trained with BPTT, which only requires storing hidden state across all time steps for efficient backprop, making complexity $\mathcal{O}(NH)$.

\section{Datasets \& Experimental Setup}
\label{sec:datasets}
\subsection{Datasets}
We consider three word/sub-word level language modeling benchmarks of different sizes. In order to make our results comparable to previous published research, we use standard datasets that have been used in prior language modeling efforts \cite{baevski2018adaptive, radford2018improving, lample2019large} - Wikitext-103 \cite{merity2016pointer}, BookCorpus \cite{zhu2015aligning} \& CC-news \cite{liu2019roberta}. We use the standard train/validation/test splits for Wikitext-103 and use random splits for the Bookcorpus and CC-news. Dataset statistics are reported in Table \ref{tab:data_stats}. Following \citet{baevski2018adaptive}, we BPE tokenize \cite{sennrich2015neural} Wikitext-103 with 32k replacements, but report perplexities at the word level for comparison with previous work, by re-normalizing the average perplexity by the actual number of words in valid/test sets. We use 30k replacements for the Bookcorpus and CC-news datasets, and report the average sub-word level perplexities since all models being compared are trained with the same tokenization and BPE pre-processing.

\begin{table}[t]
\begin{center}
\scriptsize
\begin{tabular}{|c|c|c|c|c|c|}
\midrule
\thead{Dataset} & \thead{Size \\ (GB)} & \thead{Train \\ Tokens} & \thead{Valid \\ Tokens} & \thead{Test \\ Tokens} & \thead{Vocab \\ Size} \\ 
\midrule
Wikitext-103 & 544M & 111M & 231K & 261K & 33,346 \\
Toronto BookCorpus & 3.6G & 832M & 108M & 106M & 31,300 \\
CC-news & 83G & 16.6B & 370M & 370M & 63,724 \\
\midrule
\end{tabular}
\end{center}
\caption{Dataset statistics}
\label{tab:data_stats}
\end{table}

\subsection{Experimental Details}
Across all datasets, our primary point of comparison is with a vanilla transformer language model \cite{radford2018improving}. For the Wikitext-103 and Bookcorpus datasets, we use a model that has either 12, 14 or 16 layers, $d_{model}$ = 768, 12 attention heads, $d_{ff}$ = 3,072, dropout of 0.1 everywhere including the attention scores and GeLU activations \cite{hendrycks2016gaussian} - a configuration similar to the smallest GPT-2 model \cite{radford2019language}. For the CC-News dataset, we increase $d_{model}$ to 1,024 and $d_{ff}$ to 4,096. We use a batch size of 256 randomly sampled chunks of 512 tokens. For our CC-News models, we use gradient accumulation to simulate a batch size of 256. At inference, we consider the entire piece of text as one contiguous block and use a sliding window of size 512 tokens with a stride of 256 and make predictions only over the last 256 tokens in the window. This ensures that for every minibatch of examples, the model has some context to work with.

Our multi-scale architectures use the same configuration above for every transformer layer, with added upsampling and downsampling modules. Our best-performing models use average pooling for downsampling and we present an ablation study with strided convolutions in Table \ref{tab:ablation}. We used a single transpose convolution layer with an appropriate kernel size and stride for upsampling. Each scale increases by a factor of 4 from the previous.

All models were trained with mixed precision arithmetic and optimized using Adam \cite{kingma2014adam} with the learning rate increased linearly to $2.5\times e^{-4}$ over 40,000 warmup steps and then annealed to 0 over a million steps using a cosine  schedule \cite{radford2018improving}. We tuned the dropout rates in our Wikitext-103 in the range of 0.1 to 0.5 with increments of 0.05, since we found that our models were able to overfit the data quite easily. For BookCorpus and CC-News, we only experimented with the number of transformer layers at each hierarchy and the number of hierarchies (see Table \ref{tab:ablation}).

\section{Results \& Discussion}
\label{sec:Results}

\subsection{Perplexity vs Memory Footprint trade-off}
\begin{figure}[t]
    \center{\includegraphics[scale=0.35]
    {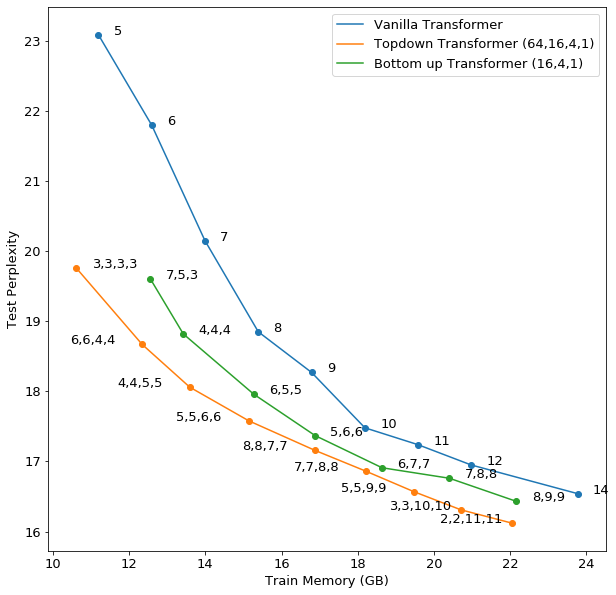}}
    \caption{\label{fig:ppl_vs_mem_book} Test perplexity vs train Memory footprint for vanilla transformer LMs and our Top-down (64x,16x,4x,1x) and Bottom-up (16x,4x,1x) multi-scale variants on the Bookcorpus test set. Numbers next to each point indicates the number of layers per scale.}
\end{figure}

As \citet{kaplan2020scaling} note, transformer language model perplexities scale as a power-law with model capacity and dataset size. Given a big enough dataset, we would therefore like to fit as big a model as our hardware will allow. Our multi-scale architectures brings us closer to this goal - demonstrating better perplexities for the same model memory footprint, as Figure~\ref{fig:ppl_vs_mem_book} shows. As expected, perplexity decreases with depth and memory usage, but the multi-scale transformer achieves lower perplexity for the  same memory usage. 
Gains are particularly stark when comparing to vanilla transformer LMs with fewer than 8 layers. 

While it is possible to manually estimate the amount of memory a model uses as shown in  Appendix Table \ref{tab:memory_footprint}, here and in the remainder of the section we use PyTorch functions\footnote{torch.cuda.max\_memory\_allocated} to report actual measurements. 

\begin{table}[t]
\begin{center}
\scriptsize
\begin{tabular}{|c|c|c|c|}
\midrule
\thead{Model} & \thead{Layers} & \thead{Test \\ PPL} & \thead{Train Mem \\ (GB)} \\
\midrule
Vanilla & 10 & 20.99 & 16.76 \\
Vanilla & 12 & 20.23 & 19.04 \\
Vanilla & 14 & 19.62 & 21.31 \\
\midrule
Top-down & 5,5,6,6 & 20.92 & 14.66 \\
Top-down & 5,5,9,9 & 20.26 & 18.35 \\
Top-down & 2,4,8,12 & 19.47 & 20.96 \\
\midrule
\end{tabular}
\end{center}
\caption{Model performance on CC-news.}
\label{tab:cc_news_ppl}
\end{table}

\begin{table}[!t]
\begin{center}
\scriptsize
\begin{tabular}{|c|c|c|c|}
\hline
\thead{Model} & \thead{Layers} & \thead{Test \\ PPL} & \thead{Mem \\ (GB)} \\
\hline
\multicolumn{4}{|c|}{\textbf{Previous Work}} \\
\hline
Vanilla \cite{welleck2019neural} & 16 & 25.6 & - \\
Transformer-XL \cite{dai2019transformer} & 16 & 24.3 & - \\
DEQ-Transformer \cite{bai2019deep} & - & 24.2 & - \\
DEQ-Transformer (Adaptive) \cite{bai2019deep} & - & 23.2 & - \\
Adaptive Inputs \cite{baevski2018adaptive} & 16 & 18.7 & - \\
\hline
\multicolumn{4}{|c|}{\textbf{Our Models}} \\
\hline
Vanilla & 16 & 25.9 & 26.85 \\
Top-down (4x, 1x) & 6,12 & 25.6 & 25.15 \\
Top-down (4x, 1x) & 5,10 & 26.1 & 22.08 \\
Bottom-up (4x, 1x) & 5,10 & 26.8 & 21.83 \\
Retina (16x, 4x, 1x) & 16 & 26.6 & 29.92 \\
\hline
\end{tabular}
\end{center}
\caption{Model performance on Wikitext-103.}
\label{tab:wiki_ppl}
\end{table}

Tables~\ref{tab:cc_news_ppl} and~\ref{tab:wiki_ppl} report perplexities of different models with their associated memory footprint. In all cases, multi-scale transformers achieve either the same perplexity as strong baselines, while using less memory; or for the same memory, they achieve lower perplexity.
On Wikitext-103, multi-scale transformers overfit the data much faster than vanilla transformers, since they have a lot more capacity for the same memory footprint. In the future, we plan to investigate better regularization methods to further improve generalization.
In Appendix Figures \ref{fig:nll_vs_word_freq_wiki} and \ref{fig:nll_vs_word_freq_book}, we further analyze which tokens are responsible for the overall improved performance similarly to~\citet{baevski2018adaptive}, and we found that the multi-scale transformer outperforms the baseline by modeling rare tokens better.

Further, in Appendix Table \ref{tab:nearest_neibs}, we report nearest neighbors in the wikitext-103 dataset using representations produced by the model at different scales, finding that retrieved chunks are often very similar in topic. In Appendix Table \ref{tab:coarse_lm}, we also report example completions from a model trained only at different coarse scales (see Appendix Section \ref{sec:coarse-only} for details).

\subsection{Analyzing Transformer LM behavior to context perturbations}
\citet{khandelwal2018sharp} and \citet{sankar2019neural} analyzed how LSTM language models and dialog systems use context, by perturbing it with different kinds of noise and observing changes in the likelihood assigned by the model to every subsequent (unperturbed) token in the sequence. \citet{Khandelwal-arxiv} show that perturbations which destroy word order, like shuffling, affects the likelihood that the model assigns to words near the shuffled context, but not far away.

\begin{figure}[t]
    \center{\includegraphics[scale=0.4]
    {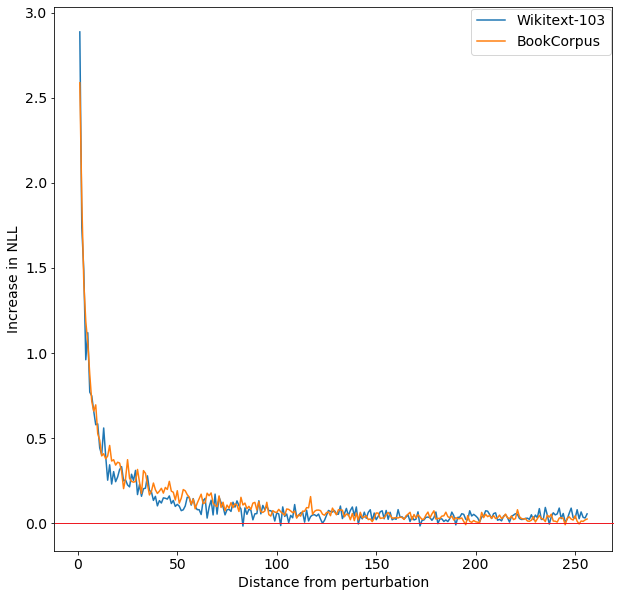}}
    \caption{\label{fig:shuffle_delta_ppl} Increase in NLL for a trained model on the Wikitext-103 and BookCorpus datasets when word order is destroyed by shuffling its context, as a function of the distance from the shuffled context.}
\end{figure}

We observe similar patterns for transformer language models on the Wikitext-103 and Bookcorpus datasets, as shown in Figure~\ref{fig:shuffle_delta_ppl}. In particular, we consider a sequence of 512 tokens, shuffle the first 256 tokens and observe the likelihood that an already trained model (without perturbations) assigns to the subsequent 256 tokens. When the shuffled context is more than 50 tokens away, the change in likelihood, compared to when the model is presented with a completely unshuffled context is already minimal, indicating that the model may not be using word order beyond this distance. This observation might partly explain why our Retina model, which only looks at fine-grained context in a local way while using coarser representations for distant context, can perform similarly as a model that uses the entire context at a fine-grained scale (see results in Table~\ref{tab:ablation}).

\subsection{Ablations}
\label{sec:ablation}
In Table~\ref{tab:ablation} we extensively investigate how the main design choices of multi-scale tranformers, such as the number of scales and number of layers at each scale, impact perplexity.

First, we observe that adding a coarser scale to the top-down multi-scale transformer helps when moving from (4,1) to (16,4,1), as these have the same memory footprint but the latter model has lower perplexity. However, we do not observe gains when adding an additional coarser scale with the (64,16,4,1) transformer.

Second, we find that the best layer allocation places more layers at the finer scales. Therefore, the designer of the architecture has to strike a trade-off between minimizing perplexity and limiting memory consumption since the finest scale is responsible for the bulk of memory usage.

Third, one may wonder whether improved perplexity reported by multi-scale models is merely due to the depth increase of the overall architecture. We thus trained skinnier but deeper vanilla transformers with $d_{model}$ set to 256 and 512 and $d_{ff}$ set to 1024, 2048, and found that they perform worse. Therefore, the gains are not entirely due to depth increase.

Fourth, in the previous section we have already discussed how transformers are not so sensitive to the order of the tokens in their farther context, motivating the use of the Retina variant of the multi-scale architecture.
Here we show a) that the performance of the baseline model only mildly deteriorates when reducing the size of the context below 64 tokens (in fact, only 3 points are lost when using a context size of 8 tokens), and b) that this property can be leveraged by the Retina version of our model. 

In particular, the ``attention context window'' column shows the context size in the multi-headed attention of the retina model. For example, "0:512" indicates that a token at position $t$ and can look at tokens $t$ to $t-512$ and "0:8, 8:256, 256:512" indicates that the attention heads at the 1x scale can look at tokens $t$ to $t-8$, $t-8$ to $t-256$ at the 4x scale, and $t-256$ to $t-512$ at the 16x scale. Using this configuration and for the same memory footprint, our Retina model is able to improve over a model that looks at all 512 words at the 1x scale.

Finally, we experimented with three different downsampling modules - average pooling, max-pooling and strided convolutions. We found average pooling to be better than convolutions and max-pooling to be quite unstable during training.

\subsection{Sample based evaluation}
A hallmark of the recent progress in language modeling has been improvements not only in perplexity, but sample quality as well. While evaluating generative models of text purely based on their samples is extremely difficult and an on ongoing effort \cite{cifka2018eval,semeniuta2018accurate}, we would like to ensure that we aren't losing out on sample quality with respect to vanilla transformer LMs.
Table \ref{tab:sample_level_eval} in the Appendix shows no significant difference in terms of generation quality between multiscale models and baseline transformers according to several common metrics; see Table \ref{tab:book_samples} in Appendix for some qualitative comparisons.

\begin{table}[ht!]
\begin{center}
\scriptsize
\begin{tabular}{|c|p{0.8cm}|p{1.3cm}|p{0.8cm}|c|p{0.6cm}|}
\hline
\thead{Model} & \thead{Scales} & \thead{Attention \\ Context \\ Window} & \thead{Layers} & \thead{Test \\ PPL} & \thead{Mem \\ (GB)} \\
\hline
\multicolumn{6}{|c|}{\textbf{Ablation for number of scales}} \\
\hline
Vanilla &  1         & 0:512 & 12       & 16.95 & 20.98 \\
Vanilla &  1         & 0:512 & 14       & 16.54 & 23.78 \\
Top-down &  4,1       & 0:512 & 8,12     & 15.87 & 25.45 \\
Top-down &  16,4,1    & 0:512 & 4,8,12   & 15.76 & 25.23 \\
Top-down &  64,16,4,1 & 0:512 & 2,4,8,12 & 16.00 & 22.62 \\
\hline
\multicolumn{6}{|c|}{\textbf{Ablation for capacity at different scales}} \\
\hline
Top-down &  64,16,4,1 & 0:512 & 16,12,6,1  & 22.14 & 10.61 \\
Top-down &  64,16,4,1 & 0:512 & 10,10,8,5  & 17.89 & 13.59 \\
Top-down &  64,16,4,1 & 0:512 & 8,8,7,7    & 17.16 & 16.87 \\
Top-down &  64,16,4,1 & 0:512 & 7,7,8,8    & 16.86 & 18.22 \\
Top-down &  64,16,4,1 & 0:512 & 5,5,9,9    & 16.57 & 19.47 \\
Top-down &  64,16,4,1 & 0:512 & 3,3,10,10  & 16.31 & 20.71 \\
Top-down &  64,16,4,1 & 0:512 & 2,2,11,11  & 16.12 & 22.05 \\
Top-down &  64,16,4,1 & 0:512 & 1,1,12,12  & 15.95 & 23.40 \\
\hline
\multicolumn{6}{|c|}{\textbf{Ablation for downsampler}} \\
\hline
\makecell{Top-down \\ (Avg-Pool)} &  64,16,4,1 & 0:512 & 7,7,8,8  & 16.86 & 18.22 \\
\makecell{Top-down \\ (Conv)}     &  64,16,4,1 & 0:512 & 7,7,8,8  & 17.95 & 18.45 \\
\hline
\multicolumn{6}{|c|}{\textbf{Ablation for deep \& narrow networks}} \\
\hline
\makecell{Vanilla \\ (256/1024)} & 1 & 0:512 & 30 & 22.15 & 22.07 \\ 
\makecell{Vanilla \\ (512/2048)} & 1 & 0:512 & 20 & 17.70 & 22.77 \\
\hline
\multicolumn{6}{|c|}{\textbf{Ablation for attention context window}} \\
\hline
Vanilla & 1 & 0:512 & 12 & 16.95 & 23.78 \\
Vanilla & 1 & 0:256 & 12 & 17.00 & 23.78 \\
Vanilla & 1 & 0:128 & 12 & 17.39 & 23.78 \\
Vanilla & 1 & 0:64  & 12 & 17.88 & 23.78 \\
Vanilla & 1 & 0:16  & 12 & 19.01 & 23.78 \\
Vanilla & 1 & 0:8   & 12 & 20.19 & 23.78 \\
\hline
Retina & 16,4,1 & 0:8,8:256,\newline256:512    & 12 & 16.81 & 23.95 \\
Retina & 16,4,1 & 0:16,16:256,\newline256:512  & 12 & 17.07 & 23.95 \\
Retina & 16,4,1 & 0:128,64:256,\newline128:512 & 12 & 17.08 & 23.95 \\
\hline
\end{tabular}
\end{center}
\caption{Model ablations for the number of scales, number of transformer layers per scale, type of downsampling function, skinny and deep networks and different local attention/retina attention masks. All reported results are on our BookCorpus test set. The attention context window column indicates what attention heads at each scale look at - for example, "0:512" implies all scales see the entire history while "0:8,8:256,256:512" indicates that the finest scale sees only the previous 8 tokens, the subsequent scale from 8-256 and so on.}
\label{tab:ablation}
\end{table}

\section{Conclusion}
We propose three multi-scale transformer architectures for language modeling and show that they achieve competitive or better perplexities compared to vanilla transformer language models for the same model memory footprint across three different language modeling benchmarks. 

These models leverage the robustness of transformers to word ordering in distant context windows. In these architectures, the representation produced by coarser scales which operate on much shorter sequences, is combined with the representation of the finest scale, thereby reducing the overall computational and memory cost.

Future work will explore how to combine our multi-scale architectures with adaptive attention head spans and how different types of information (e.g., topic, grammatical correctness) are differently affected by architecture scale choices.

\bibliography{example_paper}
\bibliographystyle{icml2020}

\clearpage

\section*{Appendix}

\subsection{NLL vs Word Frequency}
\begin{figure}[h!]
    \center{\includegraphics[scale=0.3]
    {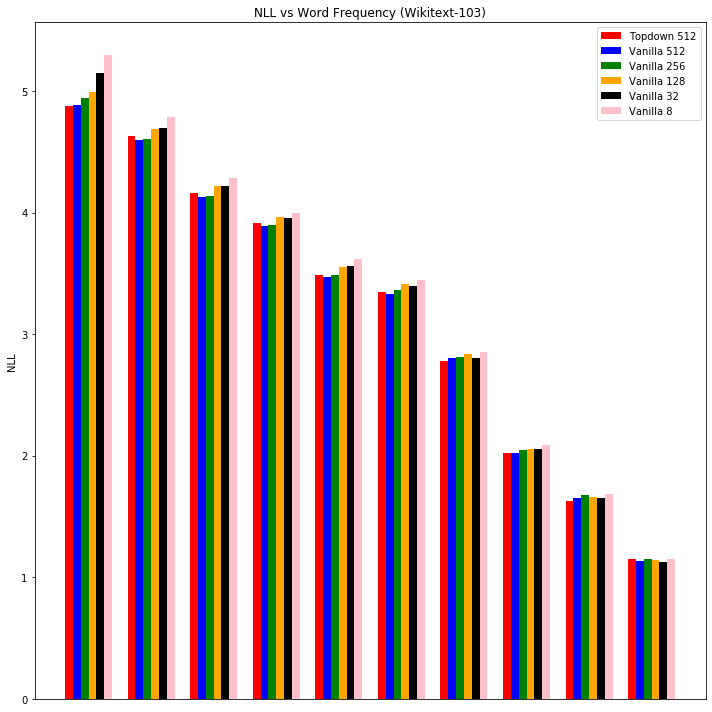}}
    \caption{\label{fig:nll_vs_word_freq_wiki} Test NLL vs word frequency on the Wikitext-103 dataset. Left to right: rarest to most frequent word bins.}
\end{figure}

\begin{figure}[h!]
    \center{\includegraphics[scale=0.3]
    {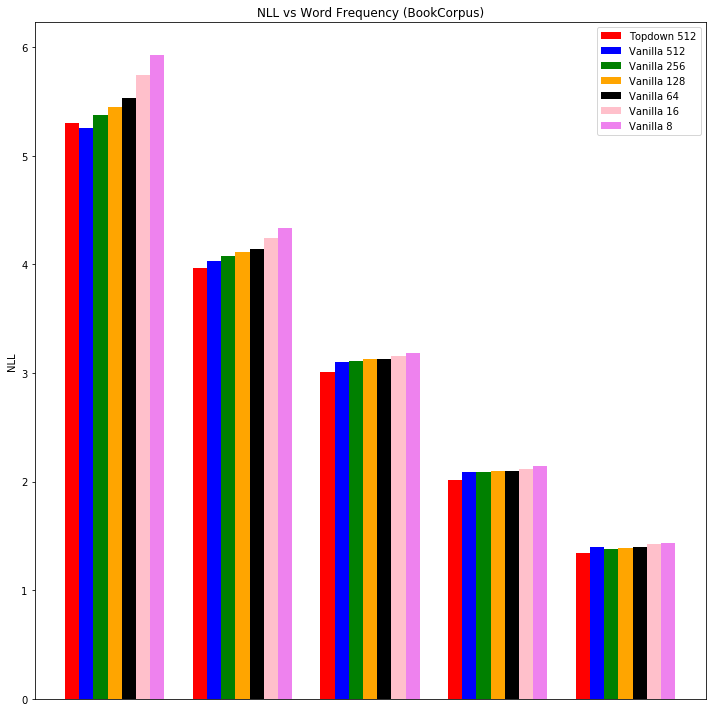}}
    \caption{\label{fig:nll_vs_word_freq_book} Test NLL vs word frequency on the BookCorpus dataset. Left to right: rarest to most frequent word bins.}
\end{figure}

In tables \ref{fig:nll_vs_word_freq_wiki} and \ref{fig:nll_vs_word_freq_book}, we break down the token-level perplexities based on their frequency of occurrence in the training set, similar to \cite{baevski2018adaptive}. We bin tokens into 5 or 10 equally sized bins, based on cumulative frequency and report the average NLL within each bin, with bins sorted from those that contain least to most frequent words. We find that the performance gap between models with smaller and larger context window sizes stems from modeling of low-frequency tokens.

\subsection{Coarse-only Transformer LMs}
\label{sec:coarse-only}
We explore the possibility of training a model that makes predictions directly at a particular scale. For example, if we consider a scale of 4, we initially downsample the input by a factor of 4 by averaging the embeddings of every non-overlapping 4-gram chunk. We then run transformer layers on this input and use the final representations at each step to predict the next \textit{4-gram} chunk. We train the model to make these predictions by minimizing the cross-entropy between the model's output distribution and a uniform distribution over the subsequent 4-gram.

Table \ref{tab:coarse_lm} presents qualitative results showing model predictions for the following. We experimented with adding this criterion of being able to predict the subsequent bag of words distribution in our top-down model at all scales, but didn't observe any improvements. This experiment mostly served as a sanity check to ensure that the representations learned at different scales can be useful. To ensure that representations at different scales learn meaningful representations, we qualitatively analyze them by looking at nearest neighbors computed using them in Table \ref{tab:nearest_neibs}. We found that nearest neighbors computed on Wikitext-103 typically returned segments within highly related topics.

\subsection{Sample-Level Evaluation Metrics}
A hallmark of the recent progress in language modeling has been improvements not only in perplexity, but sample quality as well. While evaluating generative models of text purely based on their samples is extremely difficult and an ongoing effort \cite{cifka2018eval,semeniuta2018accurate}, we would like to ensure that we aren't losing out on sample quality with respect to vanilla transformer LMs.

We therefore borrow a few sample-level evaluation metrics typically used to evaluate GAN-based text generation methods to compare samples from our model with vanilla transformer LMs. In all sample-based evaluation setups, we provide models with 64 tokens of context and generate sequences of length 256 with topk sampling \cite{fan2018hierarchical} and a temperature of 0.7.
\paragraph{N-gram and GPT-2 PPL} - We compute the likelihood of model generated samples under a pre-trained kneser-ney smoothed 4-gram language model \cite{heafield2011kenlm} and a pre-trained GPT-2 345M parameter model \cite{radford2019language}.
\paragraph{Ref BLEU} - Given an a particular input context, we generate 3 distinct completions and use these as references to compute BLEU wrt the ground truth completion.
\paragraph{N-gram repeats} - Following \citet{welleck2019neural}, we also estimate the fraction of 1,2,3 and 4-grams that the model repeats.

Quantitative results for different models are presented in Table \ref{tab:sample_level_eval}, with qualitative samples in Table \ref{tab:book_samples}. We could not observe any significant difference in sample quality between the different models. We also report the time it takes for each model to generate a sequence of 256 tokens given a context of 64 tokens.

\begin{table}[h!]
\begin{center}
\tiny
\begin{tabular}{|c|c|c|c|c|c|c|}
\midrule
\thead{Model} & \thead{Layers} & \thead{N-gram \\ PPL} & \thead{GPT-2 \\ PPL} & \thead{Ref \\ BLEU} & \thead{N-gram \\ Repeat} & \thead{Time \\ (s/seq)} \\
\midrule
Vanilla & 14 & 22.81 & 7.30 & 7.45 & 0.75/0.45/0.29/0.20 & 1.35 \\ 
\midrule
Top-down & 2,4,8,12 & 21.17 & 7.69 & 6.80 & 0.73/0.41/0.26/0.16 & 0.93  \\ 
\midrule
Bottom-up & 8,9,9 & 24.54 & 8.53 & 7.04 & 0.75/0.45/0.28/0.18 & 1.36 \\ 
\midrule
Retina & 12 & 20.91 & 8.64 & 7.37 & 0.73/0.42/0.26/0.17 & 1.72 \\
\midrule
\end{tabular}
\end{center}
\caption{Sample based evaluation of models on the BookCorpus. The N-gram repeat column reports 1/2/3/4-gram repeat fractions with a within of 256 tokens}
\label{tab:sample_level_eval}
\end{table}

\begin{table*}[h!]
\tiny
\begin{center}
\small
\begin{tabular}{|p{16cm}|}
\midrule
\textbf{Reference Text :}  \\
\midrule
good on Megadeth 's claim to being the world 's state @-@ of @-@ the @-@ art speed metal band " . Musicologist Glenn Pillsbury stated the guitar work on the album was a mixture of Mustaine 's " controlled chaos " and the " technical brilliance " of Marty Friedman . Studio efforts released in the mid- and late 1990s featured songs with compact structures and less complicated riffing . Megadeth 's lyrics often focus on death , war , politics , and religion . The lyricism centers on nihilistic themes , but occasionally deals with topics such as alienation and social problems . The earliest releases featured themes such as occultism , graphic violence \\
\midrule
\textbf{Nearest Neighbors at the 4x Scale} \\
\midrule
vocalist Jock Cheese and keyboardist / vocalist Eugene de la Hot Croix Bun , and enjoyed a large underground / independent following . Their third album , Machiavelli and the Four Seasons , reached the Australian national top 10 in 1995 . TISM were known for their hybrid of dance music and rock 'n'roll , high @-@ energy live shows and humorous lyrics . TISM 's songs frequently satirised modern culture , celebrities and the entertainment industry , classic literature and art , current affairs , politics and sport . The titles of their songs were often wordplays created by juxtaposing pop culture references with more intellectual ones ( for \\
\midrule
n ; 1950 's blues artists Guitar Slim , Johnny Watson , and B.B. King ; R \& B and doo @-@ wop groups ( particularly local <unk> groups ) ; and modern jazz . His own heterogeneous ethnic background , and the diverse social and cultural mix in and around greater Los Angeles , were crucial in the formation of Zappa as a practitioner of underground music and of his later distrustful and openly critical attitude towards " mainstream " social , political and musical movements . He frequently lampooned musical fads like psychedelia , rock opera and disco . Television also exerted a strong influence , as demonstrated by quotations from show \\
\midrule
magazine , " I am a young adult now , and I think this album shows my growth vocally . " Aaliyah was mastered by Bernie Grundman at his studio in Los Angeles . = = Music and lyrics = = An R \& B and neo soul album , Aaliyah featured midtempo funk songs , hip hop @-@ textured uptempo tracks , and slow jams that draw on older soul influences . Along with contemporary urban sounds , its music incorporated Middle @-@ Eastern influences , muted alternative rock , and , particularly on Timbaland 's songs for the album , Latin timbres . " Never No More " mixed both older soul and modern hip hop sounds with \\
\midrule
\textbf{Nearest Neighbors at the 16x Scale} \\
\midrule
well received by both critics and fans , and was responsible for bringing Slayer to the attention of a mainstream metal audience . Kerrang ! magazine described the record as " the heaviest album of all " . Alongside Anthrax 's Among the Living , Megadeth 's Peace Sells ... but Who 's Buying ? and Metallica 's Master of Puppets , Reign in Blood helped define the sound of the emerging US thrash metal scene in the mid @-@ 1980s , and has remained influential subsequently . Reign in Blood 's release was delayed because of concerns regarding its graphic artwork and lyrical subject matter . The opening track , " Angel of Death " , which refers to Josef \\
\midrule
music industry in 1992 , through his vocal contributions on Dr. Dre 's The Chronic . That album is considered to have " transformed the entire sound of West Coast rap " by its development of what later became known as the " G @-@ funk " sound . The Chronic expanded gangsta rap with profanity , anti @-@ authoritarian lyrics and multi @-@ layered samples taken from 1970 's P @-@ Funk records . Snoop Dogg contributed vocals to Dre 's solo single , " Deep Cover " , which led to a high degree of anticipation among hip hop for the release of his own solo album . Doggystyle and The Chronic are associated with \\
\midrule
the most innovative popular musicians in America if not the world " but also " the most politically ambitious . Not even in the heyday of [ the ] Clash has any group come so close to the elusive and perhaps ridiculous ' 60s rock ideal of raising political consciousness with music . " Their music on the album inspired leftist and Afrocentric ideals among rap listeners who were previously exposed to more materialist themes in the music . Reeves said it introduced black consciousness to the " hip @-@ hop youth " of the " post @-@ black power generation " , " as leather African medallions made popular by rappers like P.E. replaced thick gold chains as \\
\midrule
\textbf{Neighbors at the 64x Scale} \\
\midrule
an album that established the concept for Metallica 's following two records . Colin Larkin , writing in the Encyclopedia of Popular Music , singled out " For Whom the Bell Tolls " as an example of Metallica 's growing music potential . Popoff regards Ride the Lightning as an album where " extreme metal became art " . Megaforce initially printed 75 @,@ 000 copies of the album for the US market , while Music for Nations took care of the European market . By the autumn of 1984 , Ride the Lightning had moved 85 @,@ 000 copies in Europe , resulting in Metallica 's first cover story for British rock magazine Kerrang ! in its December issue \\
\midrule
best record of derisive punk rock since Exile on Main St. ( 1972 ) by the Rolling Stones . In The New Yorker , Ellen Willis wrote that she learned to appreciate Too Much Too Soon more than New York Dolls after seeing the band perform songs from the former album in concert , particularly " Human Being " and " Puss ' n ' Boots " , while Ron Ross from Phonograph Record magazine said the group 's " easy going ironic sensibility " was expressed " far more amusingly and accessibly " here than on their debut album . Some reviewers were critical of Too Much Too Soon for what they felt was a poorly recorded and overproduced \\
\midrule
= Riot Act features a diverse sound , including folk @-@ based and experimental songs . Stephen Thomas Erlewine of AllMusic said " Riot Act is the album that Pearl Jam has been wanting to make since Vitalogy - a muscular art rock record , one that still hits hard but that 's filled with ragged edges and odd detours . " Gossard said " Riot Act really seems to showcase all of our thing . There 's the simple rock songs we could have written in the earlier era , but it covers all the different times and dynamics we 've had and still holds together . " The musical experiments also lead several songs on the album to use alternate \\
\hline
\end{tabular}
\end{center}
\caption{Nearest neighbors computed using representations obtained at different scales}
\label{tab:nearest_neibs}
\end{table*}
\begin{table*}[h!]
\tiny
\begin{center}
\small
\begin{tabular}{|p{16cm}|}
\midrule

\textbf{Context}  \\
\midrule
Prisoner of Azkaban was the third film in the series . Radcliffe 's performance was panned by New York Times journalist A. O. Scott , who wrote that Watson had to carry him with her performance . Next was Harry Potter and the Goblet of Fire in 2005 . The film was the second @-@ highest grossing Harry Potter \\
\midrule
\textbf{2x Completions } - | film | , | \\
\midrule
\textbf{4x Completions} -  | film | the | , | and | \\
\midrule
\textbf{8x Completions} -  | the | film | of | , | . | in | and | series |\\
\midrule
\textbf{16x Completions} - | grossing | the | , | worldwide | and | \$ | in | . | million | of | @-@ | highest | grossed | film | Part | @.@ | \\
\midrule
\textbf{64x Completions} - | . | Harry | the | , | and | of | in | a | " | film | was | to | 's | @-@ | The | that | Watson | Gob@@ | Potter | for | rint | Rowling | her | as | Prison@@ | performance | with | Columbus | an | on | by | she | it | Stone | had | Times | Phoenix | at | Best | Radcliffe | Half | In | from | book | series | first | grossing | his | one | role | instal@@ | Philosop@@ | but | be | \$ | which | ) | he | ( | ab@@ | is | novel | character | \\
\midrule
\textbf{Context}  \\
\midrule
set fire to the ship to prevent her from falling into enemy hands . Patriots in small boats sailed out to the burning ship , fired some of its cannons at the British ships , took what stores and loot they could , and retreated shortly before the ship 's powder magazine exploded . = = Aftermath = = The British did not \\
\midrule
\textbf{2x Completions} - | the | have | \\
\midrule
\textbf{4x Completions} - | the | their | to | fire | \\
\midrule
\textbf{8x Completions} - | the | , | to | and | any | of | not | until | \\
\midrule
\textbf{16x Completions} - | the | , | . | of | and | to | British | a | was | in | were | that | The | 's | as | on |\\
\midrule
\textbf{64x Completions} - | . | the | , | and | of | to | British | ships | a | in | was | on | had | that | The | were | 's | for | with | by | ship | at | from | wounded | as | fleet | sailed | American | his | her | which | they | but | HMS | not | been | an | @-@ | two | after | men | returned | York | battle | harbor | be | fire | = | she | their | killed | Boston | French | it | damage | off | captured | one | guns | In | crew | into | he | schooner |\\
\midrule
\end{tabular}
\end{center}
\caption{Examples of coarse LM completions. Given a }
\label{tab:coarse_lm}
\end{table*}

\lstinputlisting[float=*t, caption={top-down Model}]{topdown.py}
\lstinputlisting[float=*t, caption={Bottom-up Model}]{bottomup.py}
\lstinputlisting[float=*t, caption={Retina Model}]{retina.py}

\begin{table*}[h!]
\begin{center}
\scriptsize
\begin{tabular}{|c|c|c|c|c|c|c|c|c|c|c|}
\midrule
\thead{Scale} & \thead{Layers} & \thead{Emb} & \thead{$Q$,$K$,$V$ \\ Proj} & \thead{$QK^T$} & \thead{$\frac{QK^T}{\sqrt{d_k}}V$} & \thead{FC} &	\thead{LN, Drop \\ Residual} & \thead{Output \\ + Grad} & \thead{Hierarchical \\ Representations} & \thead{Total \\ (GB)} \\
\midrule
\multicolumn{11}{|c|}{\textbf{Memory footprint breakdown for a single transformer LM layer at different scales}} \\
\midrule
1x  &	1 &	- &	0.150	& 0.06000	& 0.100	& 0.500	& 0.500	& - & -	& 1.325 \\
4x  &	1 &	- &	0.038	& 0.00400	& 0.025	& 0.120	& 0.120	& - & -	& 0.319 \\
16x &	1 & - &	0.009	& 0.00030	& 0.006	& 0.031	& 0.031	& - & -	& 0.079 \\
64x &	1 &	- &	0.002	& 0.00005	& 0.002	& 0.008	& 0.008	& - & -	& 0.020 \\
\midrule
\multicolumn{11}{|c|}{\textbf{Memory footprint breakdown for a 12 layer vanilla transformer}} \\
\midrule
1x &	12 & 0.05 & 1.81 & 0.80531 & 1.208 & 6.040 & 6.644 & 4.10 & - & 21.26 \\
\hline
Model & - & - & - & - & - & - & - & - & - & 0.266 \\
Total & - & - & - & - & - & - & - & - & - & 21.52 (20.98) \\
\midrule
\multicolumn{11}{|c|}{\textbf{Memory footprint breakdown for a 30 layer top-down transformer}} \\
\midrule
1x  &	8 &	0.05 &	1.06 &	0.41	& 0.7	& 3.87	& 3.52	& 3.59	 & 0.308	 & 14.27 \\
4x  &	8 &	-	 &  0.26 &	0.025	& 0.176	& 0.87	& 0.96	& -	     & 0.175	 & 2.529 \\
16x &	7 &	-	 &  0.05 &	0.001	& 0.037	& 0.18	& 0.29	& -	     & 0.042	 & 0.545 \\
64x &	7 &	-	 &  0.01 &	0.00006	& 0.008	& 0.04	& 0.05	& -	     & 0.009	 & 0.121 \\
\hline
Model & - & - & - & - & - & - & - & - & - & 0.521 \\ 
Total & - & - & - & - & - & - & - & - & - & 17.98 (18.22) \\
\midrule
\multicolumn{11}{|c|}{\textbf{Memory footprint breakdown for a 23 layer Bottom-up transformer}} \\
\midrule
1x  &	8 &	0.05 &	1.21 &	0.536	& 0.805	& 4.02	& 4.42	& 4.10	 & 0.686 & 16.19 \\
4x  &	8 &	-	 &  0.30 &	0.033	& 0.201	& 1.00	& 1.10	& -	     & 0.013	 & 2.76 \\
16x &	7 &	-	 &  0.06 &	0.002	& 0.044	& 0.22	& 0.24	& -	     & 0.003 & 0.60 \\
\hline
Model & - & - & - & - & - & - & - & - & - & 0.427 \\
Total & - & - & - & - & - & - & - & - & - & 19.97 (20.41) \\
\midrule
\end{tabular}
\end{center}
\caption{Memory footprint breakdown within different models. The number within brackets in the "Total" rows corresponds to what PyTorch reports versus our computed numbers.}
\label{tab:memory_footprint}
\end{table*}

\begin{table*}[h!]
\tiny
\begin{center}
\small
\begin{tabular}{|p{18cm}|}
\midrule
\textbf{Retina Model} \\
\midrule
\textbf{one has the bluegrass and which one has blue grass? "Zane barely resisted the impulse to shake him again. It wouldn't help, but it might help him feel better." I'm not asking where Burns told you you were. "Ty nodded, refusing to comment on that. Zane shook his head and then} \\
\midrule
reached into his pocket to pull out his phone. He turned it on and pressed the buttons for the Miami Police Department. He had a feeling that Burns had told him about the specific questions he'd asked, and he looked over at Ty. "You ever been to Texas?" he asked. Ty shook his head and looked away, giving Zane the privacy he needed. He glanced at Ty. "I 'll be in touch," he promised, and then turned to walk away. "What?" Zane asked, surprised. "You 're pretty sure of yourself. "Zane grinned and nodded." Not even in the Marines? "Ty shook his head." Never. "Zane stared at him a moment longer, then walked over to the bar and poured himself a beer. He leaned against the wall and watched Ty walk away. He couldn't help but smile. "You don't have to go." Ty stopped and turned, looking back at \\
\midrule
\textbf{top-down Model} \\
\midrule
\textbf{I told Dominic silently, I just wigged out. Dominic's gaze met mine, his dark eyes glowing, his hair falling around his cheeks. You told him to get off. He didn't. The prick hasn't even apologized. I don 't think he's going to, I replied calmly, I don} \\
\midrule
't know what to say . Dominic's voice was stern. "I will not tolerate that kind of behavior from you." I knew I had to look at him, because he was glaring at me . He wasn't being a very nice guy, but I knew he was right. He deserved to be treated like a human being. I'd never seen him so angry. "Dominic," I said softly, trying to soothe him. "I 'm sorry. I didn't mean to make you angry." I tried to make him look at me, but he looked away. "I just don't understand why you do it. You have to know that you're not normal." He sighed and looked back at me. "I know. I 'm sorry. I just don't want to be like that." I sighed, knowing he was right. I didn't want to be like that either. "Well, I 'm going to go check \\
\midrule
\textbf{Vanilla Model} \\
\midrule
\textbf{Anna's lips twisted. "Over the past few days I've discovered that there are very few things that are impossible." "He's dead, "Morgana said, as much to convince herself as the annoying pest standing before her. "I watched him die." "You betrayed him. "Morgana's} \\
\midrule
eyes narrowed. "The king is dead, and the rest of the clan has been murdered."Anna's jaw hardened with a fierce fury."What do you want?" "I want you out of my office. Now. "Anna stiffened , but before she could react, Levet was at her side, his wings flapping furiously in panic. "Sacrebleu, I beg of you. I cannot leave my sister. "Anna glared at the tiny demon. "I know." "I can 't take you with me, "Anna snapped. Levet's wings flapped in panic. "I'm not leaving without Morgana." "You can't be leaving her." "No, I can't." "Then don't leave me alone." "I can 't do it." "You can." "No." Levet's words were swallowed by a sudden roar of anger \\
\midrule
\textbf{Bottom-up Model} \\
\midrule
\textbf{door wide and I gasped as I stepped inside. It was more like a studio apartment than a guest bedroom; it was about three times the size of our room back in Seattle. The walls were painted a surprisingly warm shade of gray, with furniture in a deep, contrasting dark cherry. The bedspread was black with intricate} \\
\midrule
patterns in shades of red, gold, and green. The floors were of polished wood and a thick layer of brown dust covered the walls. The ceiling was vaulted, giving the room a more natural light. There was a huge bay window and an easel in the corner. A large bed and a large, flat-screen TV surrounded it. The only other thing that was obvious was the bathroom. There were two stalls off to the side, the toilet and sink. There was a bathroom to the left of the bedroom, opposite a large, white, marble bathroom. There were two sinks, both full of toilet paper. I walked over to the bed and looked out the window. The door was slightly ajar, and a couple of windows had a view of the lake. \\
\midrule
\end{tabular}
\end{center}
\caption{Samples from different models.}
\label{tab:book_samples}
\end{table*}

\end{document}